\documentclass[letterpaper, 10 pt, conference]{ieeeconf}  

\IEEEoverridecommandlockouts                              

\title{\LARGE \bf
Novel Blood Pressure Waveform Reconstruction from Photoplethysmography using Cycle Generative Adversarial Networks 
}

\author{
Milad Asgari Mehrabadi$^{*,1}$, Seyed Amir Hossein Aqajari$^{1}$, Amir Hosein Afandizadeh Zargari$^{1}$, \\ Nikil Dutt$^{1,2}$, and Amir M. Rahmani$^{1,2,3}$
\thanks{ $^{1}$ Department of Electrical Engineering and Computer Science, University of California Irvine, CA 92697, USA{\tt\small (*correspondence e-mail: masgarim@uci.edu)}}
\thanks{$^{2}$ Department of Computer Science, University of California, Irvine, CA 92697, USA}
\thanks{ $^{3}$ Institute for Future Health, University of California, Irvine, CA 92697, USA}
}

\usepackage{graphicx}
\usepackage{amsmath}
\usepackage{subcaption}
\usepackage{siunitx}
\usepackage{multirow,hhline}
\usepackage{comment}

\newcommand\Tstrut{\rule{0pt}{2.6ex}} 
\newcommand\Bstrut{\rule[-1.9ex]{0pt}{0pt}}

\begin{document}

\maketitle
\thispagestyle{empty}
\pagestyle{empty}

\begin{abstract}


Continuous monitoring of blood pressure (BP) can help individuals manage their chronic diseases such as hypertension, 
requiring non-invasive measurement methods in free-living conditions.
Recent approaches fuse Photoplethysmograph (PPG) and electrocardiographic (ECG) signals using 
different machine and deep learning approaches to non-invasively estimate BP;
however, they fail to reconstruct the complete signal, leading to less accurate models. 
In this paper, we propose a cycle generative adversarial network (CycleGAN) based approach to extract a BP signal known as ambulatory blood pressure (ABP) from a clean PPG signal.
Our approach uses a cycle generative adversarial network that extends the GAN architecture for domain translation,
and outperforms state-of-the-art approaches  by up to 2$\times$ in BP estimation.


\end{abstract}

\section{INTRODUCTION}
Blood pressure (BP) monitoring is
critical for early detection of  cardiovascular diseases \cite{ettehad2016blood}. 
High blood pressure
can be the source of mortality and morbidity for the aging population \cite{brouwers2021arterial}. 
Hence, continuous monitoring of BP -- via the bio-markers systolic (SBP) and diastolic blood pressure (DBP) -- 
can help diagnose chronic severe conditions. 
Currently the established method for
measuring SBP and DBP (usually in mmHg) is to use a medical-grade cuff-based instrument,
which is neither comfortable nor feasible for continuous BP monitoring in everyday settings. 

\begin{figure}[h]
  \centering
  \includegraphics[width=0.9\linewidth]{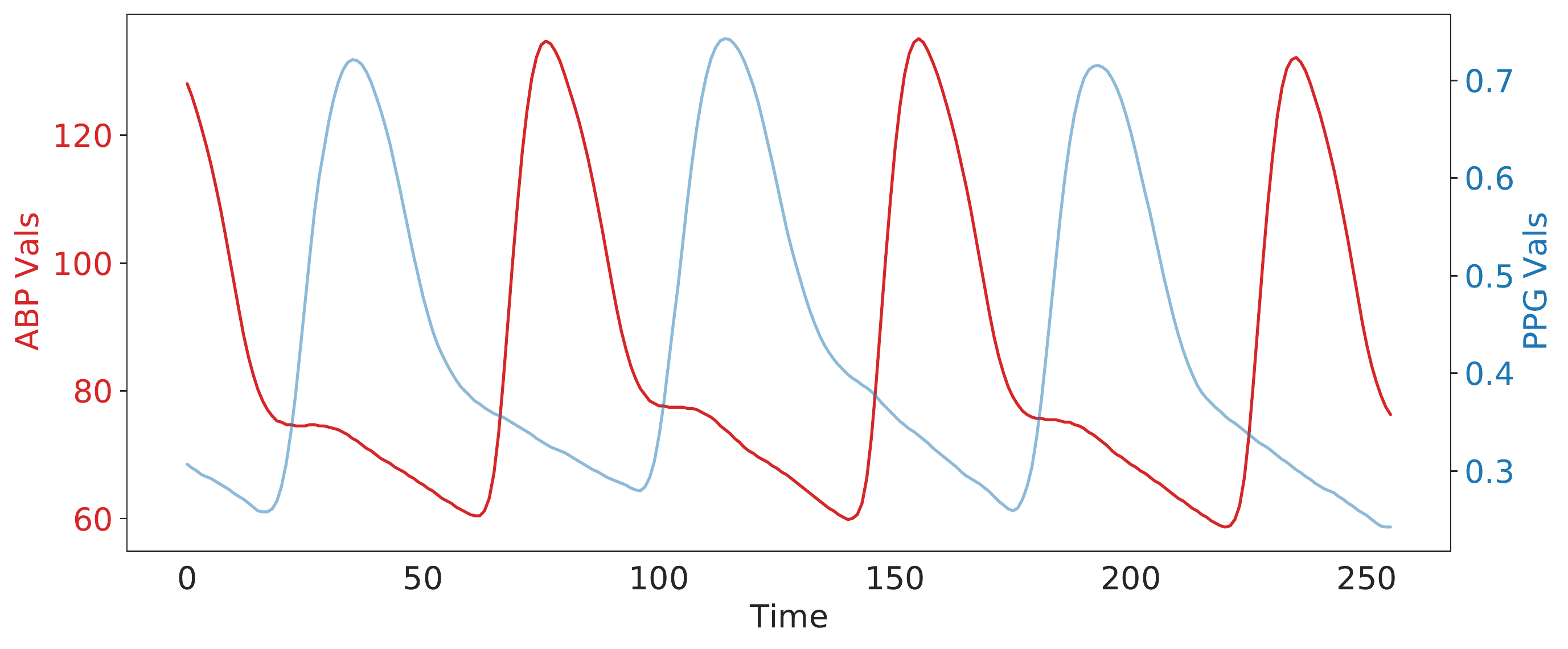}
  \caption{Correlation between BP and PPG signal.}
  \label{fig:sample}
  \vspace{-5mm}
\end{figure}

Thanks to recent advances in Internet-of-Things (IoT), it is now possible to record vital signs in everyday settings. 
For instance, smartwatches can record heart rate and blood oxygen using photoplethysmograph (PPG) and electrocardiogram (ECG) sensors. 
PPG sensors  emit and reflect light into/from blood vessels, whose measurement form a signal that is 
proportional to continuous blood volume in the unit of time.
This signal has been shown to correlate strongly with BP 
\cite{li2016novel} as shown in Fig. \ref{fig:sample}.

State-of-the-art approaches have investigated different methods ranging from feature-based statistical estimations \cite{heydari2020chest,thambiraj2020investigation,li2016novel} to deep learning-based \cite{tanveer2019cuffless,tazarv2021deep} approaches to estimate BP (i.e., SBP and DBP). 
The former approaches are limited to a pre-defined set of features that can diminish the information of the entire input signal. 
Furthermore, this needs an expert to define such features. However, the second approach utilizes convolutional layers to embed the segmented signal and absorb more information into a training model. 
Although these approaches accurately estimate SBP and DBP values, they fail to reconstruct the entire ambulatory blood pressure (ABP) signal. 
It has been shown that the waveform itself contains a rich set of information on the underlying causes of cardiovascular diseases  \cite{landry2020nonlinear,velik2015objective}. 
Moreover, to measure other cardiovascular features, such as cardiac volume, there is a need for the ABP waveform \cite{mukkamala2006continuous}.

In this paper, we propose a deep learning method based on cycle generative adversarial network (CycleGAN) \cite{zhu2017unpaired} to reconstruct the entire ABP waveform using a PPG signal.
CycleGAN has been widely used in unsupervised learning for domain transformation. 
We train and test our model on Multi-parameter Intelligent Monitoring in Intensive Care (MIMIC) II online waveform database \cite{goldberger2000physiobank} using 5-fold cross-validation with more than 90 subjects. 
In addition, since state-of-the-art approaches train and test on the same subjects' data, we perform a similar mechanism to compare our proposed model against the related work in the literature. On cross-subject evaluation, our proposed model achieves prediction error ($MAE\pm \sigma)$ of $2.89\pm 4.52$ mmHg and $3.22\pm 4.67$ mmHg for SBP and DBP, respectively. 
Furthermore, per-subject evaluation outperforms the results of the state-of-the-art methods  ($2.29\pm 0.88$ mmHg for SBP and $1.93\pm 2.61$ mmHg for DBP).
In summary, our CycleGAN method outperforms the state-of-the-art approaches for ABP waveform construction, as well as SBP and DBP estimations
improving the estimation accuracy by up to 2$\times$.


\section{Related Work}
Blood pressure estimation has been investigated in prior work \cite{heydari2020chest,tanveer2019cuffless,rong2021multi,li2016novel}. 
The majority of the related works attempt to extract SBP and DBP using ECG and/or PPG signals \cite{tazarv2021deep, tanveer2019cuffless}. 
Recent methods mainly rely on machine learning and deep learning by extracting features from the input signals \cite{mousavi2019blood}. 
The limitation of such studies would be the the under-utilization of the information in the signal. 
On the other hand, thanks to the progress in deep learning methods, researchers have focused on building deep neural networks that can generate embeddings of a given signal \cite{tazarv2021deep} to address the limitation of feature engineering. 

On the other hand, synthesizing the entire ABP signal instead of extracting numerical SBP and DBP values can lead to a valuable source of information. 
There exists a few studies on BP waveform reconstruction. 
The state-of-the-art uses statistical methods as well as a wavelet neural network to reconstruct the ABP signal \cite{li2016novel,landry2020nonlinear}. 
While these studies minimally fulfill the standards, we show the estimation accuracy can be significantly enhanced (up to 2$\times$)
by proposing a CycleGAN-based model that reconstructs the ABP signals. 

\section{Material and Methods}
\label{methods} 

\subsection{Dataset}

We employed the MIMIC-II online waveform database \cite{goldberger2000physiobank}, which contains different bio-signals of thousands of subjects hospitalized between 2001 and 2012. 
This dataset contains PPG and corresponding ABP signal with the sampling frequency of $f_s=125Hz$. 
We use 5 minutes of recording for 92 subjects to evaluate our model. 
For this dataset, we randomly select 75 subjects for training and 17 subjects for testing. 
We repeat this procedure using 5-fold cross-validation while keeping each subject's data in only one fold.
Furthermore, to compare our results with literature, we randomly select 20 subjects to train and test for each subject separate from others (the first 80\% for training and the rest for testing).

\subsection{Pre-processing}
The obtained signals had minor noises; hence, we apply the traditional Fourier Transform (FFT) approach to eliminate unwanted information. 
We use a band-pass filter with cutoff frequencies of $0.1$ and $8 Hz$ to remove noises from the PPG signal. 
On the contrary, we utilize a low-pass filter with a cutoff frequency of $5Hz$ to clean the ABP signal. For both PPG and ABP, we normalize the values of the signals. 
Afterward, each signal is divided into windows of 256 samples with 25\% overlap for the downstream learning task using CycleGAN. 

\subsection{Evaluation metrics}
We utilize the mean absolute error ($MAE$), and root mean square error ($RMSE$) to evaluate the performance of ABP construction for both SBP and DBP. 
These metrics have been widely used in the literature and show the difference between predicted and the true value. 
$MAE$ and $RMSE$ can be calculated as follows:
\begin{align}
\begin{split}
MAE_{S/DBP} = \frac{1}{N}\sum_{i=1}^{N} |T_{S/DBP}^i-P_{S/DBP}^i| \\ 
\end{split}
\end{align}

\noindent
\begin{align}
\begin{split}
RMSE_{S/DBP} = \sqrt{\frac{1}{N}\sum_{i=1}^{N}{(T_{S/DBP}^i - P_{S/DBP}^i)^2}}
\end{split}
\end{align}
As these formulas show, $MAE$ and $RMSE$ are proportional to the averages of absolute and square differences between true ($T$) and predicted ($P$) values of all samples ($N$). 

In addition, we compare our results with British Hypertension Society (BHS) \cite{o2001blood} guidelines.
This guideline divides the accuracy of blood pressure measurement into three groups based on the different ranges of estimations. 
Table \ref{tbl:bhs} summarizes these ranges with their corresponding fraction of data.

Furthermore, Bland-Altman plots have been utilized to illustrate the agreement between the true and predicted values. 
The X-axis of this plot shows the mean, and the Y-axis represents the differences between the estimated and the true value and provides 95\% confidence intervals. 

\begin{figure}[!t]
  \centering
  \includegraphics[width=0.8\linewidth]{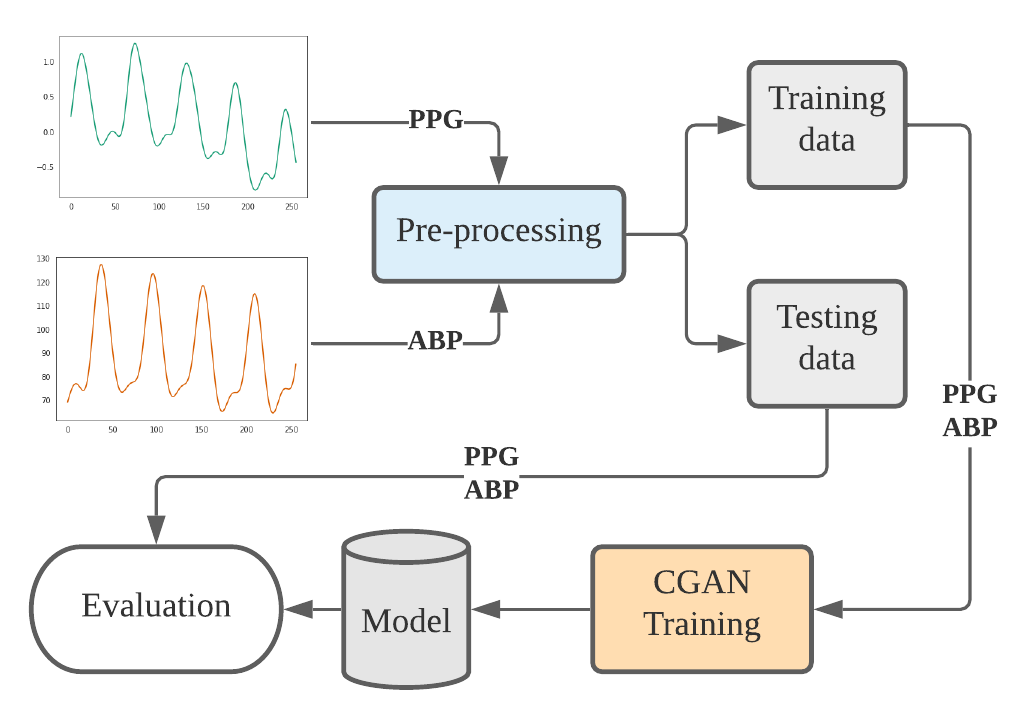}
  \caption{BP estimation pipeline.}
  \label{fig:pipeline}
  \vspace{-3mm}
\end{figure}

\begin{table}[!t]
\centering
\caption{BHS standard ranges.}
\label{tbl:bhs}
\begin{tabular}{cccc}
\hline
            & \multicolumn{3}{c}{Percentage Error}\\
            & $\le 5$ mmHg & $\le 10$ mmHg & $\le 15$ mmHg \\
\hline
\textbf{Grade A} & $60\%$ & $85\%$ & $95\%$ \\
\textbf{Grade B} & $50\%$ & $75\%$ & $90\%$ \\
\textbf{Grade C} & $40\%$ & $65\%$ & $85\%$ \\
\hline
\end{tabular}
\vspace{-4mm}
\end{table}

\subsection{PPG to ABP Translator (PAT)}

In this work, we use the Cycle Generative Adversarial Networks (CycleGAN) to reconstruct ABP signals from raw PPG signals. 
Recent studies have already shown that the CycleGAN can be employed as one of the most powerful tools for signal-to-signal translation \cite{aqajari2021end, zargari2021accurate}.

The CycleGAN proposed by Jun-Yan Zhu et al. \cite{CycleGAN2017} is an extension of the GAN architecture. The GANs are composed of a generator network and a discriminator network. 
The generator network is trained to learn the real data distribution. 
It starts from a latent space as input and attempts to generate new data similar to the original domain.
The discriminator network aims to take the generated data as an input and predict whether it is from the dataset (real) or the generated one (fake). 
After each epoch, the generator is updated to better fool the discriminator, while the discriminator is updated to accurately detect the generator's fake data.

The CycleGAN consists of two generators and two discriminator networks working in pairs. 
The idea behind the CycleGAN is to take data from the first domain as an input and generate data for the second domain as an output, and vice versa. 
In the PAT module, the goal of CycleGAN is to learn the mapping between PPG signals (domain \textit{X}) and ABP signals (domain \textit{Y}). 

Each domain contains a set of training samples $\{{x_i}\}_{i=1}^N \in X$ and $\{{y_i}\}_{i=1}^N \in Y$ used directly from MIMIC-II dataset. 
There are two generators in this module with mapping functions as $G : X \rightarrow Y$ and $F : Y \rightarrow X$. 
The two discriminators are named $D_X$ and $D_Y$. $D_X$ aims to distinguish between the real PPG signals ($x_i$) and the generated PPG signals ($F(y)$), while $D_Y$ aims to discriminate between the real ABP signals ($y_i$) and the generated ABP signals ($G(x)$).

The adversarial losses \cite{goodfellow2014generative} are used to match the distribution of the synthetic signals to the data distribution of the original signals. 
They are applied to both mapping functions ($G: X\to Y$ and $F: Y\to X$). 
The objective of the mapping function $G$ as a generator and its discriminator $D_Y$ is expressed as below: (We indicate the distributions of our data as $x \sim p_{data}(x)$ and $y \sim p_{data}(y)$.)

\noindent
\begin{align}
\begin{split}
L_{GAN}(G, D_Y, X, Y) = E_{y\sim p_{data}(y)}[log D_{Y}(y)]\\  + E_{x\sim p_{data}(x)}[log(1- D_{Y}(G(x)))]
\end{split}
\end{align}

where $G$ attempts to generate ABP signals $(G(x))$ that look similar to original ABP collected from MIMIC-II dataset (domain Y), while $D_Y$ aims to discriminate between generated ABP signals $(G(x))$ and real samples $(y)$. 
Similarly, adversarial loss for the mapping function \textit{F} is expressed as $L_{GAN}(F, D_X, Y, X)$.

The adversarial losses as the final objective loss function are not sufficient enough to guarantee that the learned functions can translate an individual input from the first domain into a desired output in the second domain. 
Therefore, cycle consistency losses are added to the final objective loss function. 
The cycle consistency losses guarantee the mapping from an individual input ($x_i$) to a desired output $(y_i)$ by considering learned mapping functions to be cycle consistent. 
Cycle consistency means for each PPG signal $x$ from domain X we must have $x \rightarrow G(x) \rightarrow F(G(x)) \approx x$ while for each ABP signals $y$ we have $y \rightarrow F(y) \rightarrow G(F(y)) \approx y$. 
The cycle consistency behavior is indicated as:

\noindent
\begin{align}
\begin{split}
L_{cyc}(G, F) = E_{x\sim p_{data}(x)}[||F(G(x))-x||_1]\\  + E_{y\sim p_{data}(y)}[||G(F(y))-y||_1]
\end{split}
\end{align}

The final objective is the weighted sum of the above loss functions:

\noindent
\begin{align}
\begin{split}
L(G, F, D_X, D_Y) = L_{GAN}(G, D_Y, X, Y) \\ + L_{GAN}(F, D_X, Y, X) \\ + \lambda L_{cyc}(G, F)
\end{split}
\end{align}

where $\lambda$ controls the relative importance of the two objective functions and is set to 10 in our work. 

$G$ aims to minimize the objective while an adversary $D$ attempts to maximize it. 
Therefore, our model aims to solve:

\noindent
\begin{align}
G^*,F^* = \arg \min_{G, F} \max_{D_X, D_Y}L(G, F, D_X, D_Y)
\end{align}

We use the CycleGAN architecture proposed by \cite{CycleGAN2017} in our work. The architecture of generative networks contains two stride-2 convolutions, nine residual blocks \cite{he2016deep}, and two fractionally-strided convolutions with stride 0.5. 
This network is adopted from Johnson et al. \cite{johnson2016perceptual}. The discriminator networks use 70$\times$70 PathGANs \cite{isola2017image} aiming to classify whether the signals are fake or real.


\section{EXPERIMENTAL RESULTS}
\label{results} 

After transforming back the signal values to the original range (due to normalization), we extract SBP and DBP and compare them with the actual ABP signal. 
State-of-the-art approaches split each subject’s data into train and test and build a separate model for each user. 
Hence, it is challenging to generalize their proposed models. 
Our study trains the entire CycleGAN on one set of subjects and tests it on new subjects given their corresponding PPG signal. 

\begin{figure}[!t]
\begin{subfigure}{.49\linewidth}
  \centering
  \includegraphics[width=\linewidth]{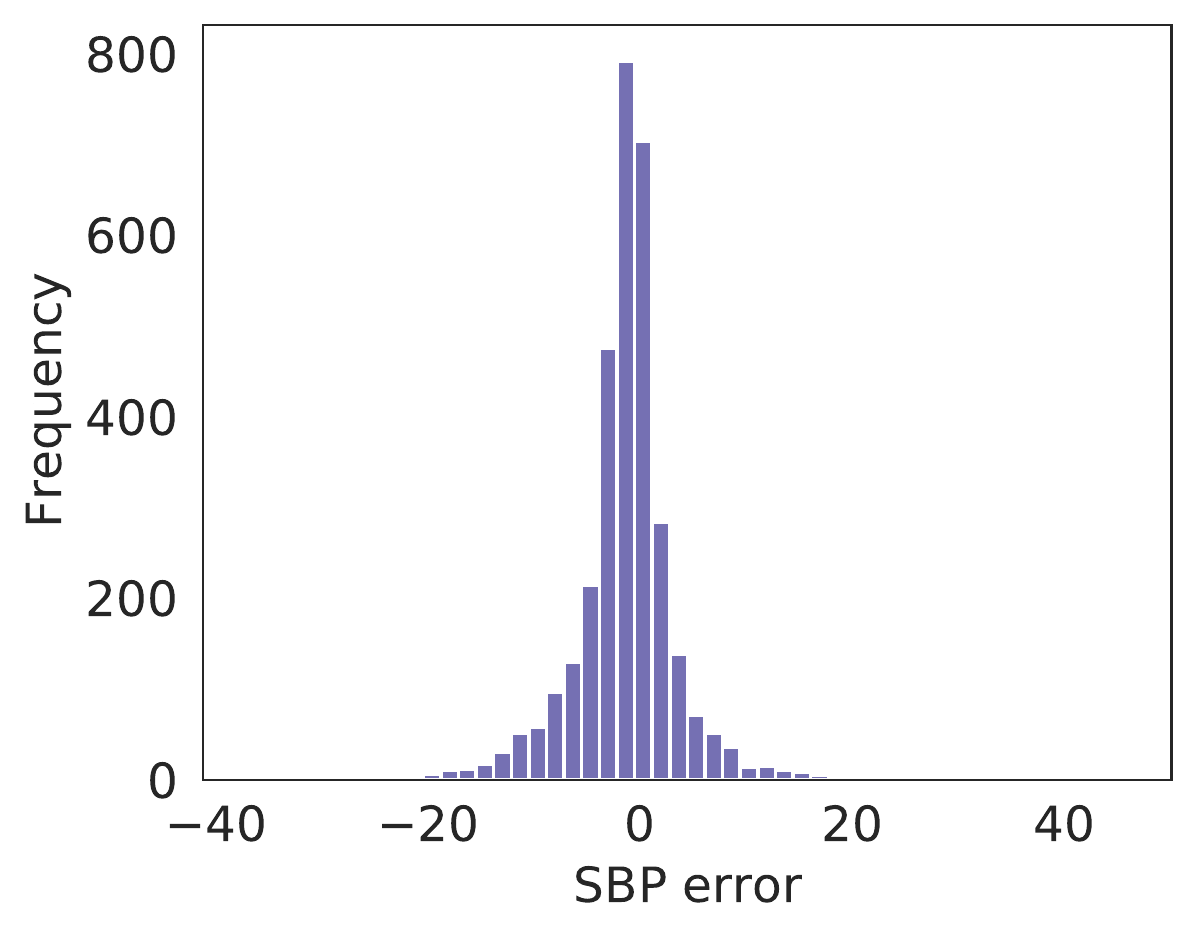}  
  \caption{}
  \label{fig:sbp_dist}
\end{subfigure}
\begin{subfigure}{.49\linewidth}
  \centering
  \includegraphics[width=\linewidth]{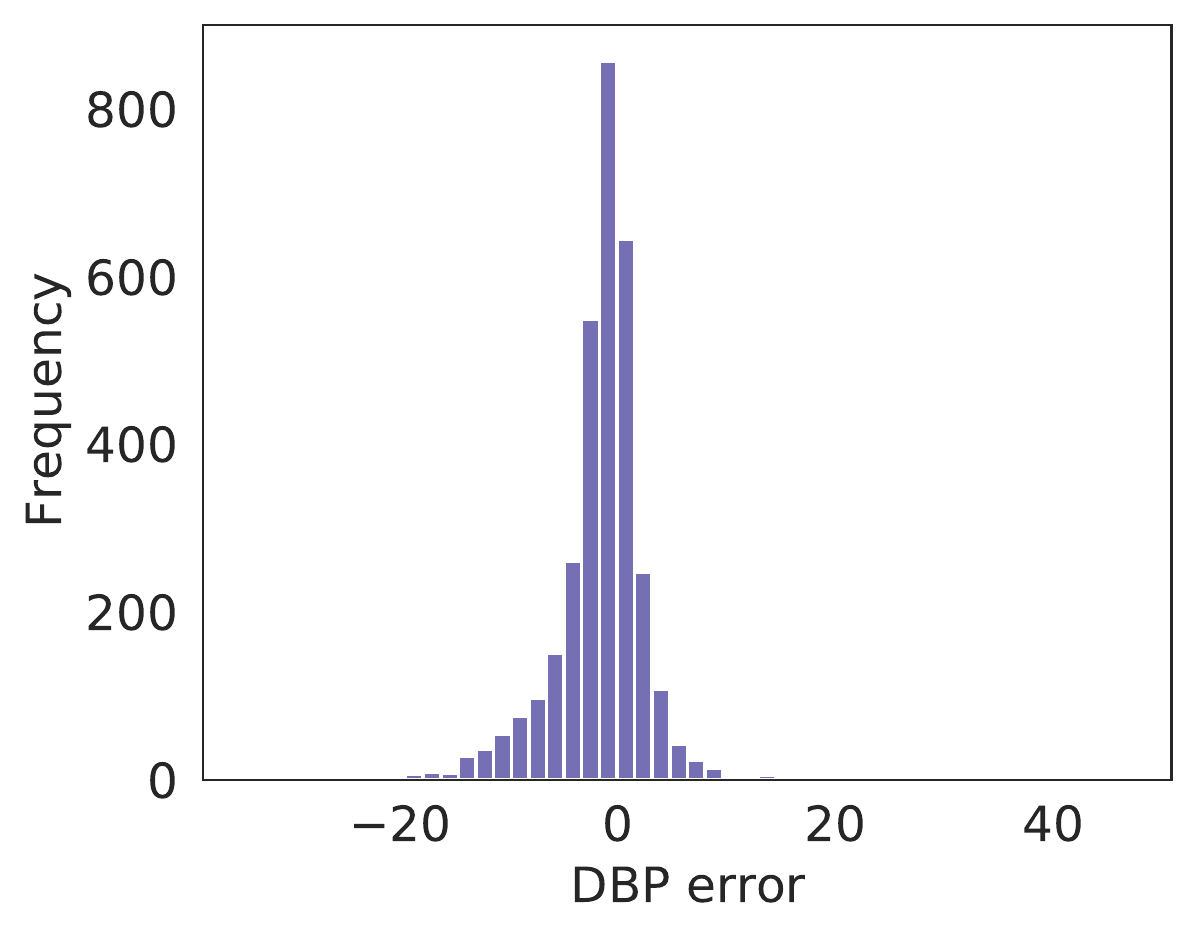}  
  \caption{}
  \label{fig:dbp_dist}
\end{subfigure}
\caption{SBP (a) and DBP (b) prediction error.}
\label{fig:dist}
\vspace{-4mm}
\end{figure}

\begin{figure}[!t]
\begin{subfigure}{.49\linewidth}
  \centering
  \includegraphics[width=\linewidth]{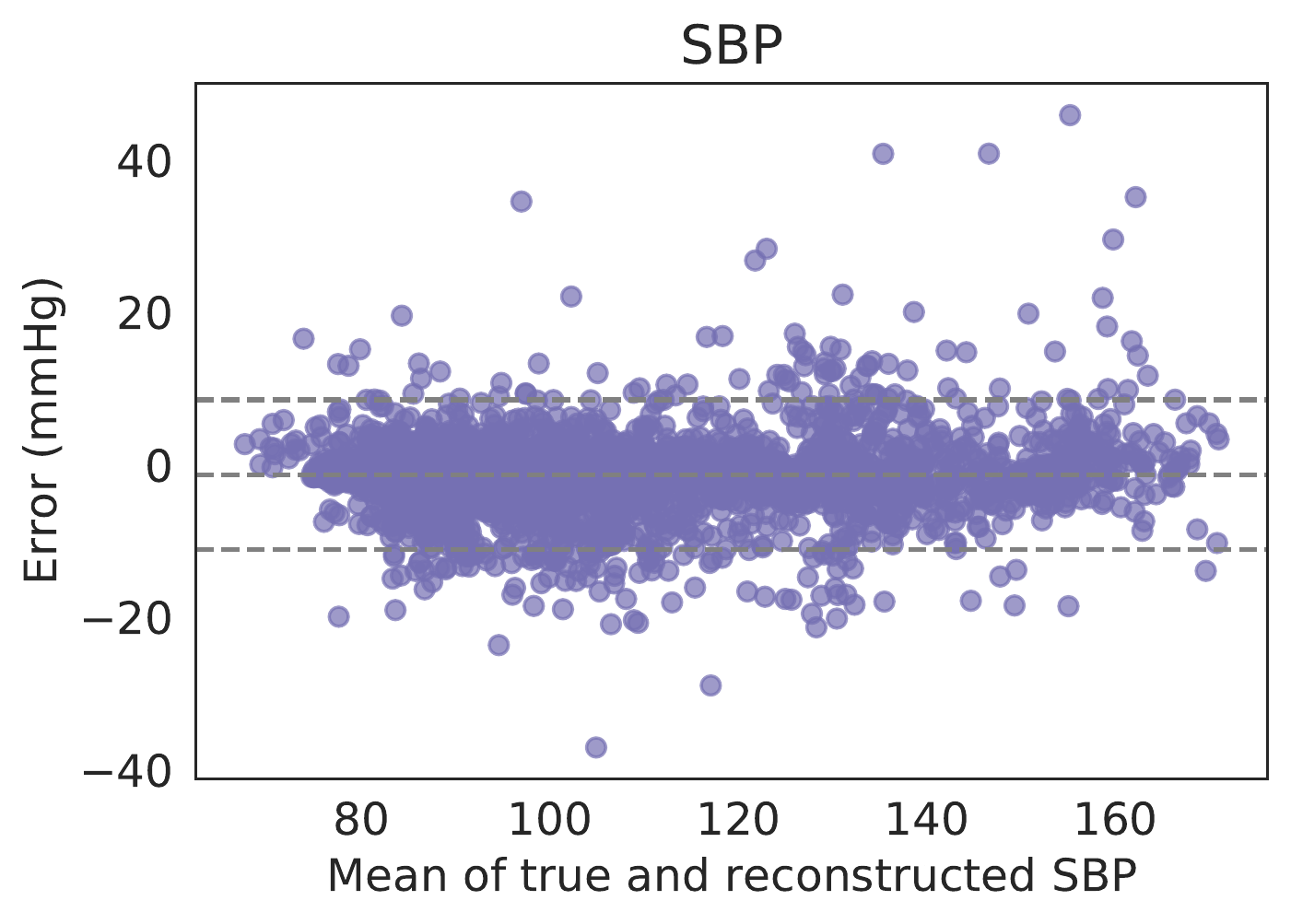}  
  \caption{}
  \label{fig:sbp_bland}
\end{subfigure}
\begin{subfigure}{.49\linewidth}
  \centering
  \includegraphics[width=\linewidth]{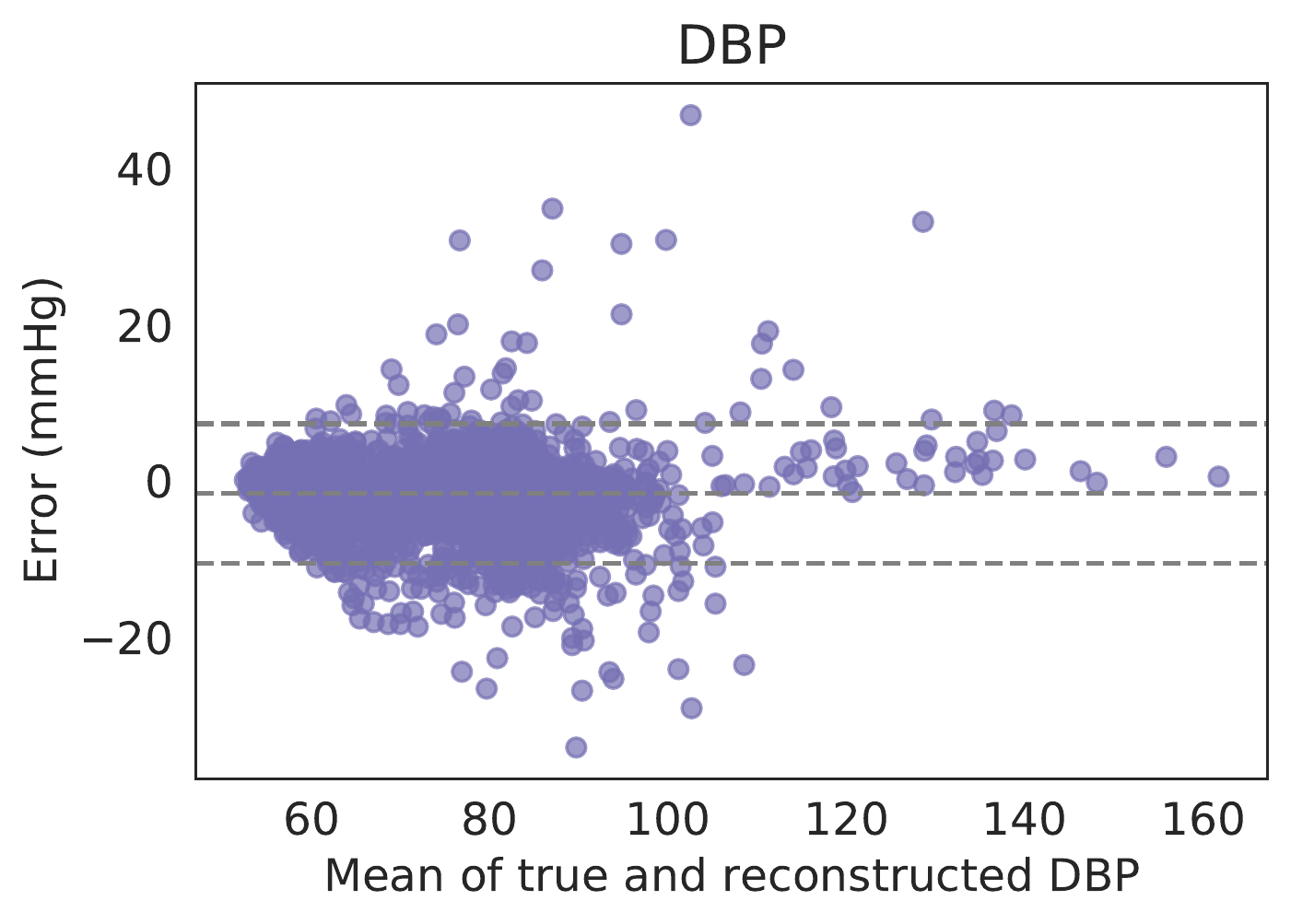}  
  \caption{}
  \label{fig:dbp_bland}
\end{subfigure}
\caption{SBP (a) and DBP (b) Bland-Altman plots.}
\label{fig:bland}
\end{figure}

\begin{table}[!t]
\centering
\caption{Average performance of our proposed CycleGAN method on MIMIC-II.}
\label{tbl:mimic2}
\begin{tabular}{ccccccc}
\hline
            & $MAE$  & $RMSE$ & $\mu$ & $\sigma$ & $r$ & ${\it P}$ \\
\hline
SBP         & 2.89 & 5.18 & 0.67 & 4.52 & 0.97 & $<.001$ \\
DBP         & 3.22 & 4.82 & 1.78 & 4.67 & 0.94 & $<.001$ \\
\hline
\end{tabular}
\vspace{-1mm}
\end{table}

Contrary to the conventional methods, we perform 5-fold cross-validation on the entire data such that each subject will be categorized to be either in the training subset or the testing subset. 
Table \ref{tbl:mimic2} summarizes the results for the MIMIC-II. 
$r$ shows Pearson correlation with the corresponding ${\it P}$-value. 
$\mu$ and $\sigma$ represent the mean and standard deviation of the estimation error, respectively.                                            
Fig. \ref{fig:dist} illustrates the prediction error distribution of SBP and DBP, respectively. Table \ref{tbl:bhs_res} summarizes our results with regards to the BHS standard. 
Our method passes the \textit{Grade A} requirements for all the criteria. 
In addition, to show how the error is distributed across different values of blood pressure, Bland-Altman plots have been utilized for both SBP and DBP (Fig. \ref{fig:bland}).

\begin{table}[!t]
\centering
\caption{Our results in regards to BHS standard ranges.}
\label{tbl:bhs_res}
\begin{tabular}{cccc}
\hline
            & \multicolumn{3}{c}{Percentage Error}\\
            & $\le 5$ mmHg & $\le 10$ mmHg & $\le 15$ mmHg \\
\hline
SBP & $85\%$ (A) & $95\%$ (A) & $98\%$ (A) \\
DBP & $81\%$ (A) & $94\%$ (A) & $98\%$ (A) \\
\hline
\end{tabular}
\end{table}

\begin{table}[!t]
\centering
\caption{Comparison of our CycleGAN-based model performance with prior works.}
\label{tbl:mimic3}
\resizebox{0.95\columnwidth}{!}{%
\begin{tabular}{ccccccccccc}
\hline
         & & \multicolumn{4}{c}{SBP}  & & \multicolumn{4}{c}{DBP} \\
\hhline{~~----~----}
         & & $MAE$ & $RMSE$ & $\mu$ & $\sigma$ & & $MAE$ & $RMSE$ & $\mu$ & $\sigma$ \\
\hline

 \multirow{3}{*}{\rotatebox[origin=c]{90}{\scriptsize Waveform}}& {\bf Our work}                 & {\bf 2.29} & {\bf 3.22} & 0.88 & 2.99 & & {\bf 1.93} & {\bf 2.61} & 0.91 & 2.32 \Tstrut\\
 & \cite{landry2020nonlinear} & 5.9  &  -    &   0.9 & 5    & & 3.5  &   -   &  0.9  & 3.5 \\
 & \cite{li2016novel}               &  -    &  -    &  2.32 & 2.91 &  & -    &   -   &  1.92 & 2.47 \Bstrut\\
\hline
\multirow{6}{*}{\rotatebox[origin=c]{90}{\parbox{1cm}{\scriptsize SBP-DBP \\ estimation}}} & \cite{huang2022mlp}    & 6.32 & 8.78 & 0.69 & 8.75 &  & 3.89 & 5.48 & 1.23 & 5.34 \\
& \cite{mousavi2019blood}  & 3.97 & 8.9  & 0.050 & 7.99 &  &2.43 & 4.18 & 0.187 & 3.37 \\
& \cite{tazarv2021deep}     & 3.70 & -    & 0.21  & 6.27 &  &2.02 & -    & 0.24 & 3.40 \\
& \cite{lin2021energy}   & 3.42 & 5.42 & 0.06  & 4.19 &  &2.21 & 3.29 & 0.18  & 2.65 \\ 
& \cite{rong2021multi}    & 3.36 &  -    &  -     & 4.48 &  &5.59 &   -   &   -    & 7.25 \\  
& \cite{kurylyak2013neural}  & 3.8  &  -    &  -     & 3.46 &  &2.21 &  -    &  -     & 2.09 \\
\hline
\end{tabular}
}
\end{table}

\begin{figure}[!t]
\begin{subfigure}{.49\linewidth}
  \centering
  \includegraphics[width=\linewidth]{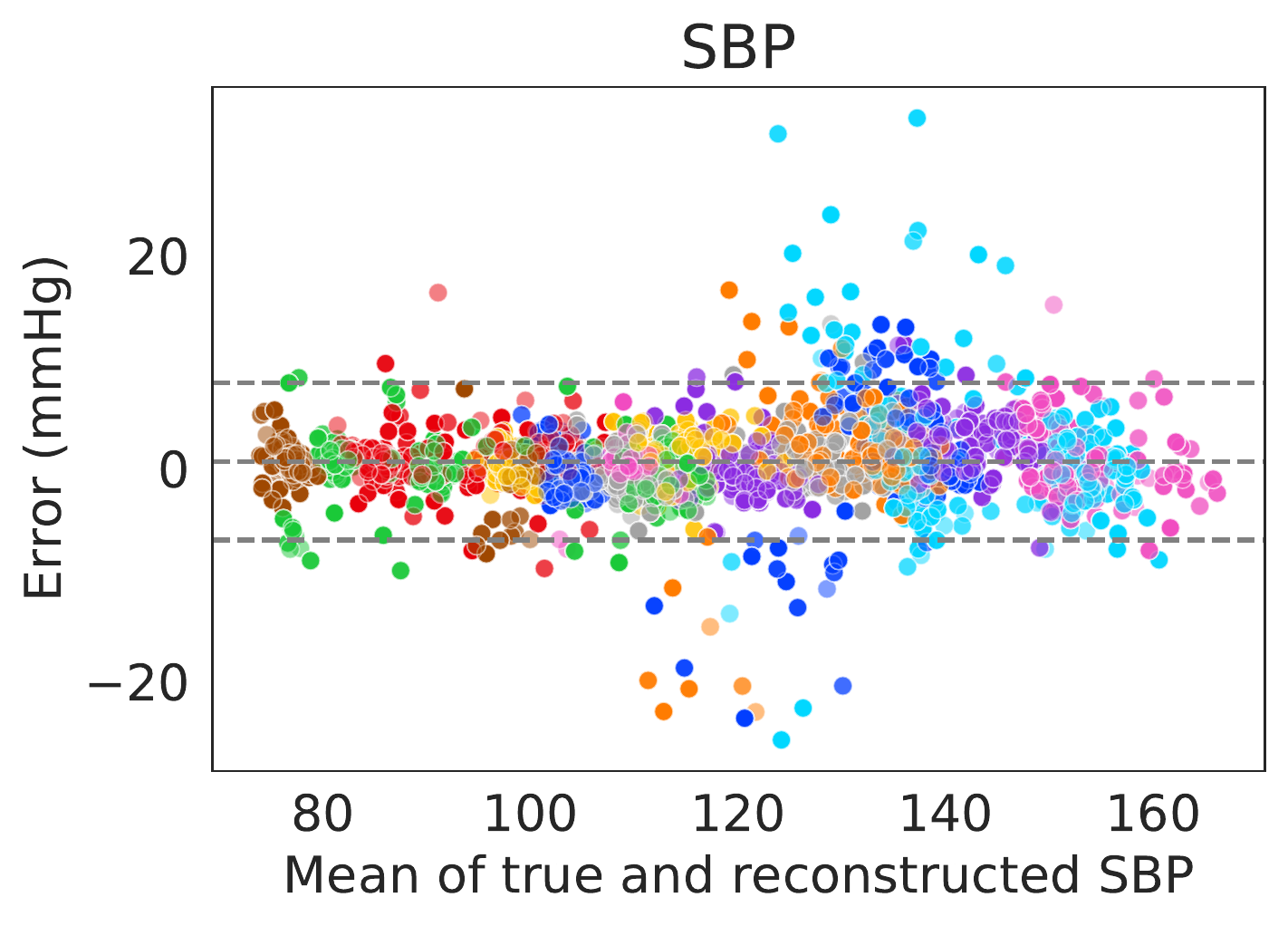}  
  \caption{}
  \label{fig:sbp_bland_per}
\end{subfigure}
\begin{subfigure}{.49\linewidth}
  \centering
  \includegraphics[width=\linewidth]{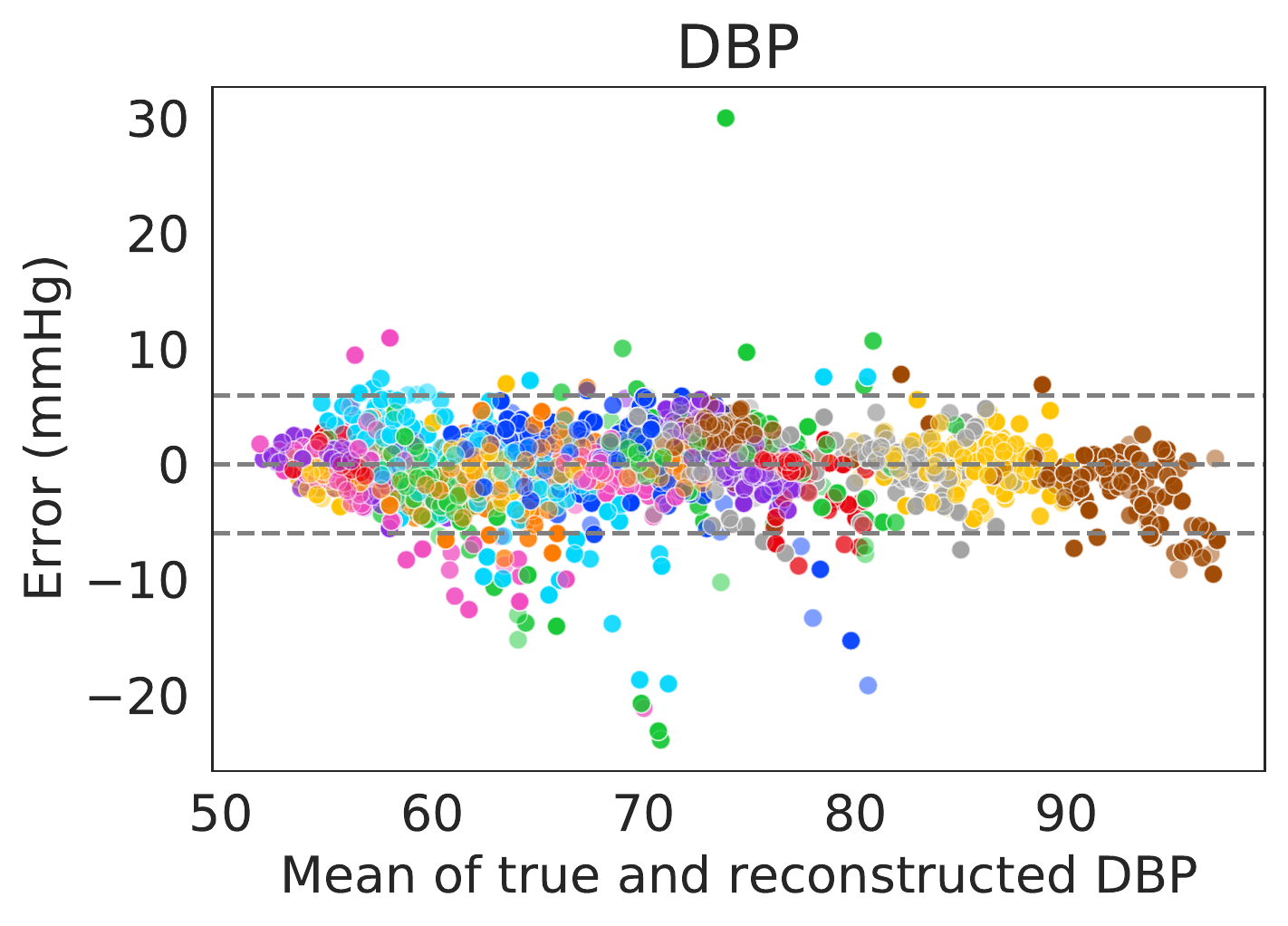}  
  \caption{}
  \label{fig:dbp_bland_per}
\end{subfigure}
\caption{Per-subject Bland-Altman plots for SBP (a) and DBP (b).}
\label{fig:bland_per}
\vspace{-5mm}
\end{figure}

 To be able to compare our results against the related work in the literature, we perform per-subject train and test procedures for a fair comparison. 
 For this purpose, we select 20 random subjects from the MIMIC-II database whose ABP and PPG signals are retained. 
 Table \ref{tbl:mimic3} shows the results of our proposed model as well as the recent related work for per-subject evaluation. Our model outperforms the studies with waveform reconstruction as well as those with only SBP-DBP values estimation.
 Moreover, Fig. \ref{fig:bland_per} presents the Bland-Altman plots for these subjects (the colors show different subjects) within the agreement limits (95\% confidence intervals).

\section{Conclusion}
\label{conclusion}
Monitoring blood pressure is essential for early detection and treatment of cardiovascular disease. 
Conventional methods utilize statistical and machine learning models to estimate SBP and DBP using PPG, a low-cost and straightforward signal; however, they fail to generate the entire signal or synthesize an accurate waveform. 
In this work, we leveraged the cycle generative adversarial network (CycleGAN) for the first time in the blood pressure estimation domain, and our model achieved $MAE$ of $2.89$ mmHg and $3.22$ mmHg for SBP and DBP in a cross-subject setting. 
The per-subject evaluation’s performance was $2.29$ mmHg for SBP and $1.93$ mmHg for DBP, outperforming state-of-the-art approaches by up to 2$\times$  for improving the BP estimation accuracy.







\bibliographystyle{IEEEtran}
\bibliography{ref}

\begin{thebibliography}{10}
\providecommand{\url}[1]{#1}
\csname url@samestyle\endcsname
\providecommand{\newblock}{\relax}
\providecommand{\bibinfo}[2]{#2}
\providecommand{\BIBentrySTDinterwordspacing}{\spaceskip=0pt\relax}
\providecommand{\BIBentryALTinterwordstretchfactor}{4}
\providecommand{\BIBentryALTinterwordspacing}{\spaceskip=\fontdimen2\font plus
\BIBentryALTinterwordstretchfactor\fontdimen3\font minus
  \fontdimen4\font\relax}
\providecommand{\BIBforeignlanguage}[2]{{%
\expandafter\ifx\csname l@#1\endcsname\relax
\typeout{** WARNING: IEEEtran.bst: No hyphenation pattern has been}%
\typeout{** loaded for the language `#1'. Using the pattern for}%
\typeout{** the default language instead.}%
\else
\language=\csname l@#1\endcsname
\fi
#2}}
\providecommand{\BIBdecl}{\relax}
\BIBdecl

\bibitem{ettehad2016blood}
D.~Ettehad \emph{et~al.}, ``Blood pressure lowering for prevention of
  cardiovascular disease and death: a systematic review and meta-analysis,''
  \emph{The Lancet}, vol. 387, no. 10022, pp. 957--967, 2016.

\bibitem{brouwers2021arterial}
S.~Brouwers \emph{et~al.}, ``Arterial hypertension,'' 2021.

\bibitem{li2016novel}
P.~Li \emph{et~al.}, ``Novel wavelet neural network algorithm for continuous
  and noninvasive dynamic estimation of blood pressure from
  photoplethysmography,'' \emph{Science China Information Sciences}, vol.~59,
  no.~4, pp. 1--10, 2016.

\bibitem{heydari2020chest}
F.~Heydari \emph{et~al.}, ``A chest-based continuous cuffless blood pressure
  method: Estimation and evaluation using multiple body sensors,''
  \emph{Information Fusion}, vol.~54, pp. 119--127, 2020.

\bibitem{thambiraj2020investigation}
G.~Thambiraj \emph{et~al.}, ``Investigation on the effect of womersley number,
  ecg and ppg features for cuff less blood pressure estimation using machine
  learning,'' \emph{Biomedical Signal Processing and Control}, vol.~60, p.
  101942, 2020.

\bibitem{tanveer2019cuffless}
M.~S. Tanveer \emph{et~al.}, ``Cuffless blood pressure estimation from
  electrocardiogram and photoplethysmogram using waveform based ann-lstm
  network,'' \emph{Biomedical Signal Processing and Control}, vol.~51, pp.
  382--392, 2019.

\bibitem{tazarv2021deep}
A.~Tazarv \emph{et~al.}, ``A deep learning approach to predict blood pressure
  from ppg signals,'' in \emph{2021 43rd Annual International Conference of the
  IEEE Engineering in Medicine \& Biology Society (EMBC)}.\hskip 1em plus 0.5em
  minus 0.4em\relax IEEE, 2021, pp. 5658--5662.

\bibitem{landry2020nonlinear}
C.~Landry \emph{et~al.}, ``Nonlinear dynamic modeling of blood pressure
  waveform: Towards an accurate cuffless monitoring system,'' \emph{IEEE
  Sensors Journal}, vol.~20, no.~10, pp. 5368--5378, 2020.

\bibitem{velik2015objective}
R.~Velik, ``An objective review of the technological developments for radial
  pulse diagnosis in traditional chinese medicine,'' \emph{European Journal of
  Integrative Medicine}, vol.~7, no.~4, pp. 321--331, 2015.

\bibitem{mukkamala2006continuous}
R.~Mukkamala \emph{et~al.}, ``Continuous cardiac output monitoring by
  peripheral blood pressure waveform analysis,'' \emph{IEEE Transactions on
  Biomedical Engineering}, vol.~53, no.~3, pp. 459--467, 2006.

\bibitem{zhu2017unpaired}
J.-Y. Zhu \emph{et~al.}, ``Unpaired image-to-image translation using
  cycle-consistent adversarial networks,'' in \emph{Proceedings of the IEEE
  international conference on computer vision}, 2017, pp. 2223--2232.

\bibitem{goldberger2000physiobank}
A.~L. Goldberger \emph{et~al.}, ``Physiobank, physiotoolkit, and physionet:
  components of a new research resource for complex physiologic signals,''
  \emph{circulation}, vol. 101, no.~23, pp. e215--e220, 2000.

\bibitem{rong2021multi}
M.~Rong \emph{et~al.}, ``A multi-type features fusion neural network for blood
  pressure prediction based on photoplethysmography,'' \emph{Biomedical Signal
  Processing and Control}, vol.~68, p. 102772, 2021.

\bibitem{mousavi2019blood}
S.~S. Mousavi \emph{et~al.}, ``Blood pressure estimation from appropriate and
  inappropriate ppg signals using a whole-based method,'' \emph{Biomedical
  Signal Processing and Control}, vol.~47, pp. 196--206, 2019.

\bibitem{o2001blood}
E.~O'brien \emph{et~al.}, ``Blood pressure measuring devices: recommendations
  of the european society of hypertension,'' \emph{Bmj}, vol. 322, no. 7285,
  pp. 531--536, 2001.

\bibitem{aqajari2021end}
S.~A.~H. Aqajari \emph{et~al.}, ``An end-to-end and accurate ppg-based
  respiratory rate estimation approach using cycle generative adversarial
  networks,'' \emph{arXiv preprint arXiv:2105.00594}, 2021.

\bibitem{zargari2021accurate}
A.~H.~A. Zargari \emph{et~al.}, ``An accurate non-accelerometer-based ppg
  motion artifact removal technique using cyclegan,'' \emph{arXiv preprint
  arXiv:2106.11512}, 2021.

\bibitem{CycleGAN2017}
J.-Y. Zhu \emph{et~al.}, ``Unpaired image-to-image translation using
  cycle-consistent adversarial networkss,'' in \emph{2017 IEEE ICCV}, 2017.

\bibitem{goodfellow2014generative}
I.~J. Goodfellow \emph{et~al.}, ``Generative adversarial networks,''
  \emph{arXiv preprint arXiv:1406.2661}, 2014.

\bibitem{he2016deep}
K.~He \emph{et~al.}, ``Deep residual learning for image recognition,'' in
  \emph{Proceedings of the IEEE CVPR}, 2016.

\bibitem{johnson2016perceptual}
J.~Johnson \emph{et~al.}, ``Perceptual losses for real-time style transfer and
  super-resolution,'' in \emph{ECCV}.\hskip 1em plus 0.5em minus 0.4em\relax
  Springer, 2016.

\bibitem{isola2017image}
P.~Isola \emph{et~al.}, ``Image-to-image translation with conditional
  adversarial networks,'' in \emph{Proceedings of the IEEE conference on CVPR},
  2017.

\bibitem{huang2022mlp}
B.~Huang \emph{et~al.}, ``Mlp-bp: A novel framework for cuffless blood pressure
  measurement with ppg and ecg signals based on mlp-mixer neural networks,''
  \emph{Biomedical Signal Processing and Control}, vol.~73, p. 103404, 2022.

\bibitem{lin2021energy}
W.~Lin \emph{et~al.}, ``Energy-efficient blood pressure monitoring based on
  single-site photoplethysmogram on wearable devices,'' in \emph{2021 43rd
  Annual International Conference of the IEEE Engineering in Medicine \&
  Biology Society (EMBC)}.\hskip 1em plus 0.5em minus 0.4em\relax IEEE, 2021,
  pp. 504--507.

\bibitem{kurylyak2013neural}
Y.~Kurylyak \emph{et~al.}, ``A neural network-based method for continuous blood
  pressure estimation from a ppg signal,'' in \emph{2013 IEEE International
  instrumentation and measurement technology conference (I2MTC)}.\hskip 1em
  plus 0.5em minus 0.4em\relax IEEE, 2013, pp. 280--283.

\end{thebibliography}

\end{document}